\documentclass[sn-apa,iicol]{sn-jnl}

\usepackage{graphicx}%
\usepackage{multirow}%
\usepackage{amsmath,amssymb,amsfonts}%
\usepackage{amsthm}%
\usepackage{mathrsfs}%
\usepackage[title]{appendix}%
\usepackage{xcolor}%
\usepackage{textcomp}%
\usepackage{manyfoot}%
\usepackage{booktabs}%
\usepackage{algorithm}%
\usepackage{algorithmicx}%
\usepackage{algpseudocode}%
\usepackage{listings}%
\usepackage{booktabs}
\usepackage{graphicx}
\usepackage{caption}
\usepackage{adjustbox}

\usepackage{booktabs}
\usepackage{xcolor}
\usepackage{amsmath}
\usepackage{amssymb}
\usepackage{array}
\usepackage{color}
\usepackage{colortbl}
\usepackage{framed}
\usepackage{bm}
\usepackage{bbm}
\usepackage{xspace}
\usepackage{enumitem}
\usepackage{xparse}
\usepackage{url}
\usepackage{wrapfig}
\usepackage{soul}

\usepackage{pythonhighlight}

\usepackage[accsupp]{axessibility}  

\definecolor{Gray}{gray}{0.6}
\definecolor{LightCyan}{rgb}{0.88,0.95,1}
\definecolor{blond}{rgb}{0.98, 0.94, 0.75}

\def \ie {\emph{i.e.}}
\def \eg {\emph{e.g.}}

\usepackage[dvipsnames]{xcolor}

\newcommand{\rev}[1]{#1}
\newenvironment{revonly}{}{}

\usepackage{wrapfig} 
\usepackage{booktabs} 
\usepackage{amsmath} 
\usepackage{amsfonts} 
\usepackage{courier}
\usepackage{multirow}
\usepackage{listings} 
\usepackage{svg}
\usepackage{pgfplots, filecontents}
\pgfplotsset{compat=1.17} 

\newcommand{\tit}[1]{\smallbreak\noindent\textbf{#1.}}
\newcommand{\tinytit}[1]{\noindent\textbf{#1.}}

\newcommand{\HEAD}{{HEaD+}\xspace}  
\newcommand{\OLDHEAD}{{HEaD}\xspace}  
\newcommand{\DATASETNAME}{{InsideGen}\xspace}  
\newcommand{\PFItc}{$\text{PFI}_{\mathcal{T}}$\xspace}  
\newcommand{\CT}{\mathcal{T}}  
\newcommand{\perfect}{\textit{complete}\xspace}
\usepackage{lipsum} 

\definecolor{commentcolor}{RGB}{110,154,155}   

\usepackage{epstopdf}

\theoremstyle{thmstyleone}%
\theoremstyle{thmstyletwo}%

\theoremstyle{thmstylethree}%

\begin{document}

\title[Hallucination Early Detection\\ in Diffusion Models]{Hallucination Early Detection in Diffusion Models}

\author*[1]{\fnm{Federico} \sur{Betti}}\email{federico.betti@unitn.it}
\equalcont{These authors contributed equally to this work.}

\author[2]{\fnm{Lorenzo} \sur{Baraldi}}\email{lorenzo.baraldi@phd-unipi.it}
\equalcont{These authors contributed equally to this work.}

\author[3]{\fnm{Lorenzo} \sur{Baraldi}}\email{lorenzo.baraldi@unimore.it}

\author[3]{\fnm{Rita} \sur{Cucchiara}}\email{rita.cucchiara@unimore.it}

\author[1]{\fnm{Nicu} \sur{Sebe}}\email{nicu.sebe@unitn.it}

\affil*[1]{\orgname{University of Trento}, \orgaddress{\country{Italy}}}

\affil[2]{\orgname{Universit\`a di Pisa}, \orgaddress{\country{Italy}}}

\affil[3]{\orgname{University of Modena and Reggio Emilia}, \orgaddress{\country{Italy}}}


\abstract{Text-to-Image generation has seen significant advancements in output realism with the advent of diffusion models.
However, diffusion models encounter difficulties when tasked with generating multiple objects, frequently resulting in hallucinations where certain entities are omitted. While existing solutions typically focus on optimizing latent representations within diffusion models, the relevance of the initial generation seed is typically underestimated.
While using various seeds in multiple iterations can improve results, this method also significantly increases time and energy costs.
To address this challenge, we introduce \HEAD (Hallucination Early Detection +), a novel approach designed to identify incorrect generations early in the diffusion process. The \HEAD framework integrates cross-attention maps and textual information with a novel input, the Predicted Final Image. The objective is to assess whether to proceed with the current generation or restart it with a different seed, thereby exploring multiple-generation seeds while conserving time. \HEAD is trained on the newly created \DATASETNAME dataset of 45,000 generated images, each containing prompts with up to seven objects. Our findings demonstrate a 6-8\% increase in the likelihood of achieving a \perfect generation (\ie~an image accurately representing all specified subjects) with four objects when applying \HEAD alongside existing models. Additionally, \HEAD reduces generation times by up to 32\% when aiming for a \perfect image, enhancing the efficiency of generating complete and accurate object representations relative to leading models.
\rev{Moreover, we propose an integrated localization module that predicts object centroid positions and verifies pairwise spatial relations (if requested by the users) at an intermediate timestep, gating generation together with object presence to further improve relation-consistent outcomes.}}


\maketitle

\section{Introduction}
\label{sec:intro}

\begin{figure*}[t]
    \centering
    \includegraphics[width=\linewidth]{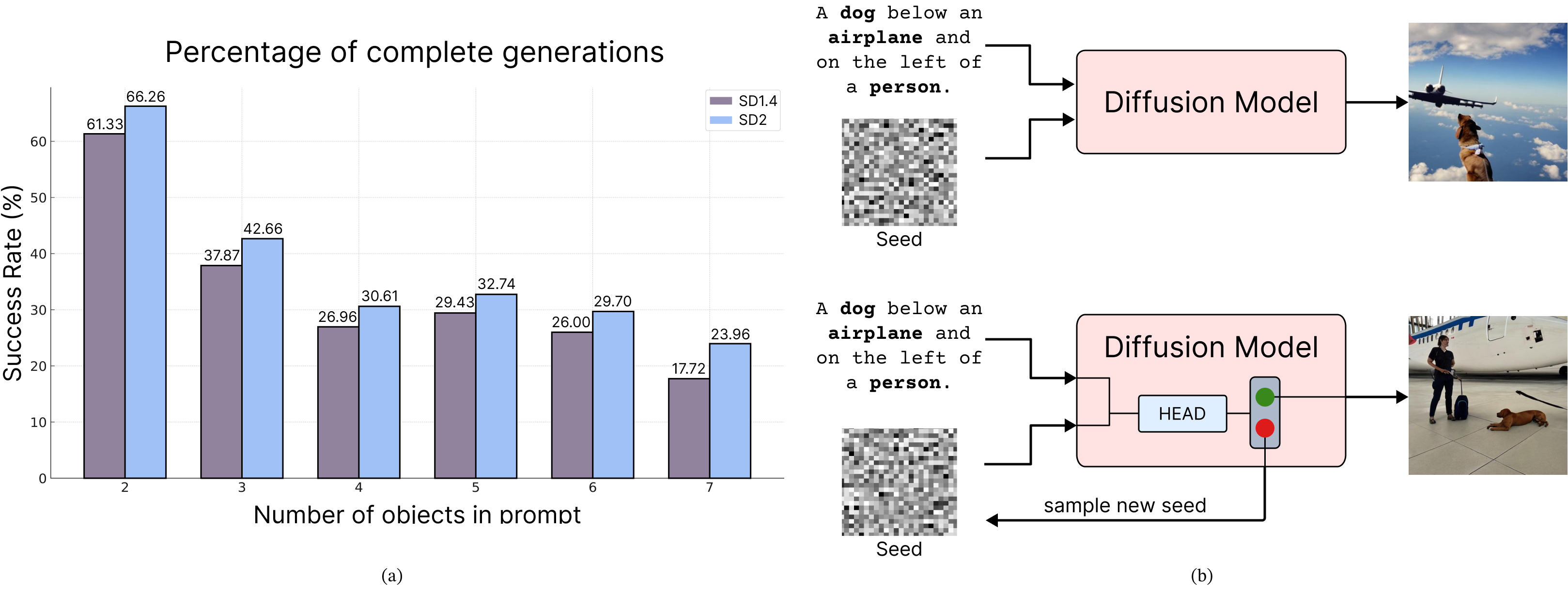}
    \caption{
a) Proportion of \textit{complete} generations, \ie~images featuring all requested objects, as a function of the number of objects requested in the prompt (\DATASETNAME dataset). b) Summary of the \HEAD pipeline: throughout the generation process, \HEAD evaluates whether all specified objects will be correctly depicted in the final image, determining whether to proceed with the current generation or restart with a different seed.
    }
    \label{fig:first_page}
    \vspace{-.3cm}
\end{figure*}

The advent of diffusion models~\citep{sohl2015deep,ho2020denoising} (DMs) and their latent variants~\citep{rombach2022high} has significantly enhanced the fidelity and alignment of generated content with user-provided context~\citep{dhariwal2021diffusion}. Nonetheless, these models continue to face challenges in faithfully representing the requested prompts. For instance, when a prompt specifies multiple objects, there is a considerable probability that not all objects will be accurately depicted, resulting in an \textit{incomplete} generation~\citep{chefer2023attendandexcite,agarwal2023star,wang2023tokencompose}.

These inaccuracies are problematic for both users, who may not receive the intended content, and for companies, as the production of erroneous images results in wasted computational resources and energy.

Furthermore, diffusion models have exhibited subpar performance in generating ``long-tail" objects~\citep{zhang2023deep}, often misrepresenting rare concepts in the generated images~\citep{samuel2023generating}. This challenge has led researchers to explore the influence of initial seeds on the generation process, investigating their effects and developing criteria for their optimal selection~\citep{samuel2023generating}. Since different seeds can produce varying results, selecting the appropriate seed could significantly mitigate both the long-tail issue and the problem of hallucinations.

In our study, we assess the performance of Stable Diffusion v1.4 (SD1.4) and Stable Diffusion v2 (SD2)~\citep{rombach2022high} when tasked with generating images from prompts that request the presence of a different number of objects. Our analysis reveals that the probability of a \perfect generation -- \ie an image showing all requested objects -- with four objects is only 26.96\% for SD1.4 and improves to 30.61\% for SD2, as illustrated in Fig. \ref{fig:first_page}(a). The effectiveness of an image generation diffusion model depends strongly on the choice of the initial seed, which establishes the starting latent noise and guides the model through the latent space~\citep{karthik2023dont, samuel2023norm, liu2023compositional, wu2023harnessing, chefer2023attendandexcite}. This aspect is underlined by our results: in scenarios with four objects, at least one among the 11 seeds we tested leads to a \perfect generation in 79.87\% of the prompts. This significant increase highlights the central role of seed selection in overcoming the unpredictability and variability associated with these models, indicating substantial opportunities for improving generative accuracy.

While some attempts have been made to develop automatic evaluation metrics for image generation~\citep{betti2023let, lu2023llmscore}, these still fail at ensuring a sufficiently fast and reliable evaluation. Further, even in the presence of a reliable assessment of the generated image, the generation process should be repeated different times until reaching a \perfect generation. In this paper, we instead take a different path and explore solutions for an early (\ie, at early stages of the diffusion process) detection of hallucination. We use this in combination with a procedure to abort or restart generation with another seed to save time and improve the final quality.
To this aim, we focus on specific hallucinations: the omission in the generated image of one or more target objects requested in the textual prompt.

Our approach, termed \textbf{\HEAD}, or Hallucination Early Detection+, is the first approach designed to enhance both the efficiency and accuracy of generative DMs. \HEAD leverages intermediate cross-attention maps and textual information to assess the connection between the input prompt and the internal attention operations of the model. Further, \HEAD operates on the \emph{Predicted Final Image} (PFI), which denotes a forecast of the final generated image at a specific time step of the generation process. The combination of PFIs and cross-attention maps allows for the early identification of potential errors by predicting the inclusion or exclusion of objects requested by the initial prompt. This proactive error identification allows \HEAD to suggest terminating the diffusion process early, thereby optimizing resource usage and reducing the time spent on generating images that are unlikely to meet quality standards.
\rev{In addition to object presence, we equip \HEAD with a localization module that, when enabled, predicts object centroid positions to verify simple pairwise relations (\eg~left/right, above/below) when specified in the prompt. Ultimately, this decision is combined with presence predictions to determine whether to continue or restart generation.}

\HEAD is developed through the training of an Hallucination Prediction (HP) network on a dataset featuring both accurate and hallucinated images. Specifically, we generate \textbf{\DATASETNAME}, a dataset that encompasses 45,000 images generated with SD1.4~\citep{rombach2022high} and SD2, saving cross-attention maps and PFIs at intermediate steps. Leveraging on this dataset, \HEAD detector is designed to integrate seamlessly with all existing DMs, enhancing their ability to reliably produce images that depict all requested objects. This improvement is evident when \HEAD is applied to SD1.4 and TokenCompose~\citep{wang2023tokencompose}, even with datasets and prompt categories that are entirely different from those in \DATASETNAME.

A preliminary exploration of hallucination detection was previously conducted in~\citep{betti2024optimizing}, where we tackled hallucinations in a more controlled scenario involving two subjects. However, when applied to a more complex experimental setting, the previous model exhibited limited efficiency, achieving only 11.55\% time savings. In contrast, this work introduces a more sophisticated dataset, \DATASETNAME, featuring complex and realistic prompts. Further, we improve model performance by changing the overall architecture of the HP network to incorporate the textual information associated with hallucinated objects and by better combining intermediate information.

\tit{Contributions} To sum up, our main contributions are as follows:
\begin{itemize}[noitemsep,topsep=0pt]
    \item We propose \HEAD, a novel framework designed to assess the presence of objects in images during the generation process. This opens the way to predicting aspects of the final output in mid-generation, with potential benefits on the entire image generation pipeline using diffusion models.
    \item Our approach is based on a new indicator for the early detection of objects in the generated images, the Predicted Final Image. Moreover, it employs an Hallucination Prediction network that combines PFI, cross-attention maps, and textual embeddings via a Transformer Decoder approach. This network is model-agnostic, \ie~it can be added to any diffusion models without the need for re-training.
    \item Our experimental findings show that using \HEAD in combination with SD1.4 or TokenCompose improves the probability of accurately generating an image from a four-object prompt by 7.85\% and 6.52\%, respectively.
    \item As a complimentary contribution, we build and release the \DATASETNAME dataset, which annotates hallucinations and records cross-attention maps and PFIs at various stages of the generation. \DATASETNAME dataset is available for download on the project page\footnote{Project page available at: \href{https://aimagelab.github.io/HEaD/}{aimagelab.github.io/HEaD}}.
\end{itemize}

\section{Related Works}
\label{sec:related}
\tit{Text-to-Image Generation Evaluation}
Evaluating the alignment between a generated image and its initial prompt remains a challenging task, with no universally accepted solutions currently available.
Among the assessment metrics, CLIPScore~\citep{hessel2021clipscore} evaluates the cosine similarity between the prompt and the image, both having undergone processing through their respective visual and textual CLIP backbones~\citep{radford2021learning}. On the other hand,~\cite{betti2023let} have introduced a novel scoring mechanism that harnesses the capabilities of Large Language Models (LLMs) and Visual Question Answering. Consistent with this research direction, several other investigations~\citep{lu2023llmscore,hu2023tifa,ku2023viescore,singh2023divide} propose diverse methodologies, framing their research within the common reasoning paradigm advanced by LLMs. Additionally, the use of Visual Question Answering in the context of text-to-image evaluation has been separately explored by~\cite{lin2024evaluating}.

While these methods have demonstrated their proficiency in highlighting the hallucinatory aspects of generative models, they still require the generated image as input, which is produced only in the final step of the diffusion process.
Moreover, they include additional processing beyond the generation phase, thus introducing delays in the overall evaluation due to the utilization of foundational models in the evaluation pipeline. In contrast, \HEAD enables the detection of hallucinations during the generative process, thereby preventing the creation of images that are inconsistent with their respective prompts.

\tit{Attention Maps for Image Generation}
The integration of attention mechanisms has been a cornerstone in improving image synthesis. Cross-attention layers~\citep{rombach2022high} have significantly improved visual fidelity, a concept further explored by~\cite{hertz2022prompttoprompt} to maintain coherence between text prompts and visual outputs. The role of semantic layouts in image synthesis has also been emphasized for quality and interpretability~\citep{wang2022semantic}. Extending these concepts,~\cite{rassin2023linguistic} tackled linguistic binding in diffusion models with their SynGen approach, which aligns attention maps with prompt syntax to improve attribute correspondence, optimizing the generation process without retraining the model.
~\cite{10.1145/3581783.3612191} introduced a novel method of controlling image synthesis by editing initial noise images, revealing that pixel blocks in initial latent images can be manipulated to influence a specific content generation.~\cite{balaji2022ediffi} proposed eDiff-I, an ensemble of expert denoisers for text-to-image diffusion models, which enhances text alignment and visual quality by specializing models for different synthesis stages.
\rev{Concurrently, PixArt-$\alpha$~\citep{chen2024pixartalpha} introduces a Transformer-based diffusion model that integrates text via cross-attention into a Diffusion Transformer (DiT)~\cite{Peebles_2023_ICCV}, employs a decomposed training strategy, and leverages dense pseudo-captioning, improving image quality up to 1024px with markedly reduced training costs.}

Cross-attention maps have been also employed for correcting missing-object problems in image generation. For instance, TokenCompose~\citep{wang2023tokencompose} proposes to fine-tune the Diffusion U-Net by enforcing consistencies between cross-attention maps and relative object segmentation maps. Attend-and-Excite~\citep{chefer2023attendandexcite} refines the latent embedding on the fly, by maximizing the cross-attention activation for the most neglected subject. Further, cross-attention maps have been found relevant for enhancing binding between represented subjects and their properties~\citep{Li2023divide}. Similarly,~\cite{agarwal2023star}, proposes an attention segregation loss to reduce the overlap of different concepts. In contrast, Structured Diffusion~\citep{feng2023trainingfree} adjusts cross-attention representations by exploiting linguistic structures, thus exhibiting enhanced proficiency in the generation of images with complex compositional semantics. Layout Guidance~\citep{chen2024training} introduces a methodology to direct the image generation process by optimizing cross-attention maps in alignment with a predefined layout. Furthermore, Composable Diffusion~\citep{liu2022compositional} adopts a novel approach by segmenting the prompt into distinct conditions and applying a score-based mechanism to refine the image quality.

Following the consensus on the effectiveness of cross-attention as a telltale sign of the fidelity of the generation, our work exploits this information as a factor for predicting the likelihood of accurate generation from early diffusion steps.

\tit{Seed Importance in Image Generation}
\label{sec:related_seed_selection}
In text-to-image generation, images are significantly impacted by the starting seed of the diffusion process.
Indeed, different seeds can produce completely different images, as highlighted by~\cite{karthik2023dont}, which claims to generate better-aligned images by evaluating multiple seeds.
Furthermore, \citep{10.1145/3581783.3612191,samuel2023norm} propose to edit the image by manipulating the initial noise instead of steering the generation process with additional mechanisms.

Seed selection has gained relevance in the generation of long-tail concepts~\citep{zhang2023deep}. As underlined
by~\cite{samuel2023generating}, in the generation of rare subjects, training predominantly involves exposure to a limited segment of the initial noisy latent space. This selective exposure during training contributes to the generation of unsatisfactory outcomes across a majority of generative seeds at inference time. Hence, the exploration of diverse generative seeds remains a critical aspect in enhancing generative outcomes.
To mitigate the occurrence of hallucinations, \HEAD suggests altering the seed in the event of detecting hallucinations in the generative process.

\section{Preliminaries}
\label{sec:preliminaries}
In this section, we provide a detailed explanation of latent diffusion models, noise scheduling, and text conditioning, providing the background on relevant definitions and notations.

\tit{Latent Diffusion Models}
In this study, we examine the Stable Diffusion (SD) architecture~\citep{rombach2022high}, which performs the diffusion process over a latent space of a variational autoencoder decoder~\citep{Kingma2014} instead of the traditional pixel image space. Initially, an encoder $E$ converts an image $x$ into a latent code $z = E(x)$.
The decoder $D$ seeks to achieve precise reconstruction, ensuring that $D(E(x)) \approx x$.
Within this framework, a denoising diffusion probabilistic model (DDPM)~\citep{ho2020denoising, sohl2015deep} operates. This model works on the latent space, creating a denoised version of the input latent $z_t$ at each timestep $t$. Significantly, the procedure is improved by the inclusion of a conditioning vector $c(y)$, usually obtained from a textual prompt $y$ using a CLIP text encoder~\citep{radford2021learning}.

The final training objective consists of minimizing the following loss function:
\begin{equation}
\label{eq:loss_function}
L = \mathbb{E}_{z \sim E(x), y, \epsilon \sim \mathcal{N}(0,1), t} \left\lVert \epsilon - \epsilon_{\theta}(z_t, t, c(y)) \right\rVert^2,
\end{equation}
where $\epsilon_{\theta}$ is a UNet network~\citep{ronneberger2015u}, with attention layers, that aims to predict the added noise $\epsilon$.

To obtain the final image from the denoised latent representation, the last step involves passing the final latent representation through a the decoder $D$. This decoder translates the latent space back into the pixel space, thus completing the image generation process. The transition from the final latent state $z_0$ to the generated image $x_0$ can thus be described by
\begin{equation}
\label{eq:vae_decoder}
x_0 = D(z_0).
\end{equation}
For further details, we refer the reader to~\citep{rombach2022high}.

\tit{Schedulers in Diffusion Models}
In diffusion models, schedulers are used to manage the denoising process, guiding image generation by adjusting noise levels throughout the steps. These algorithms facilitate the transition from a noisy latent representation to a refined image without adversarial training.
In our \HEAD approach, we have adapted the scheduler's function to extract the PFI at intermediate diffusion steps. This adjustment is designed to obtain the most precise representation of the final image during the generation process.

The transition of latents $z_t$ at time step $t$ to a subsequent state $z_{t'}$ is governed by the following procedure. The predicted noise $\epsilon_t$ is firstly estimated from the output of the UNet model, and  the updated latents $z_{t'}$ are computed through the update function of the scheduler $\Delta$, as
\begin{equation}
\begin{split}
\label{eq:scheduler}
\epsilon_t &= \epsilon_{\theta}(z_t, t) \\
z_{t'} &= \Delta(z_t, \epsilon_t, t, t').
\end{split}
\end{equation}
In this formulation, $\epsilon_t$ is derived by the current latents and time step, while $\Delta$ is the scheduler update function computing the new latents $z_{t'}$ based on the predicted noise $\epsilon_t$. The specific characteristics of $\Delta$ depend on the chosen scheduler, which ultimately determines the intricate dynamics of the denoising process.

\begin{figure*}[t]
    \centering
    \includegraphics[width=\textwidth]{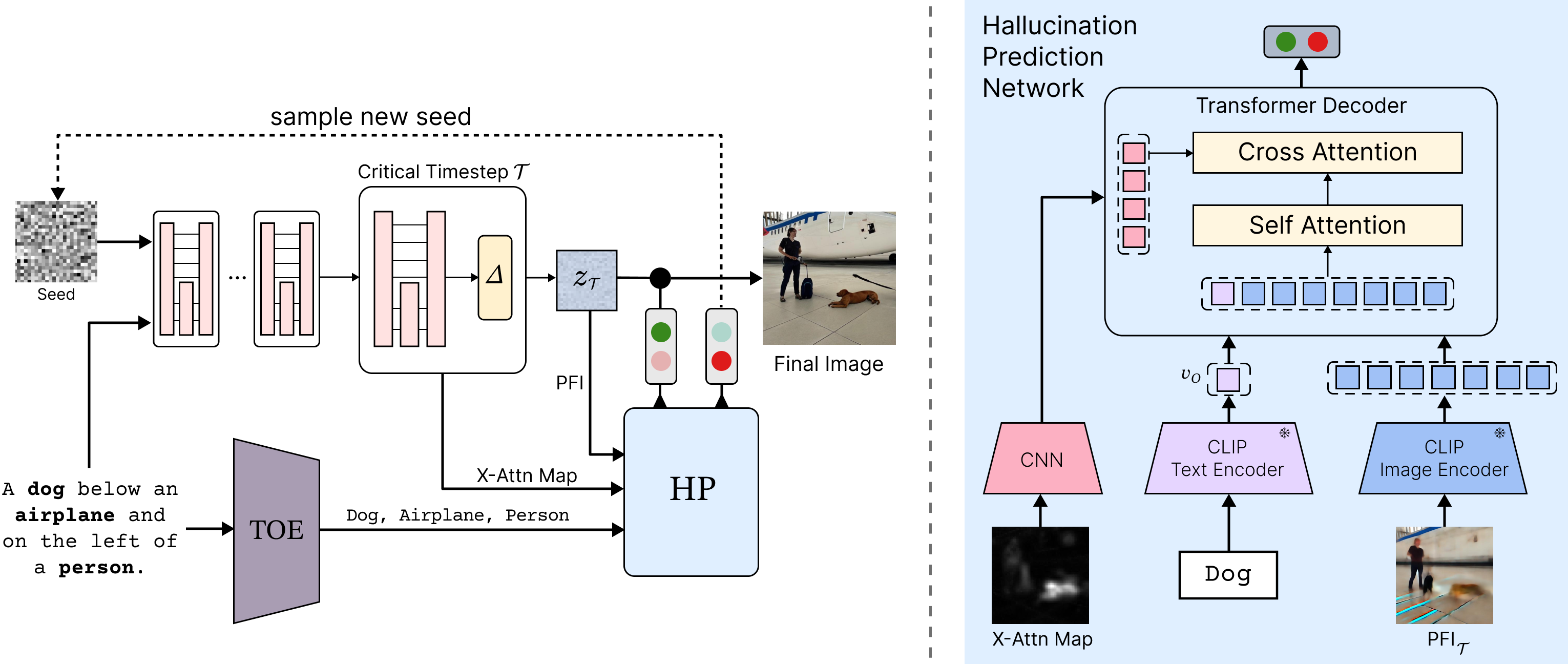}
    \vspace{-.1cm}
    \caption{Overview of the \HEAD process and detail of the Hallucination Prediction network. The process starts with extracting subjects from the prompt using the TOE model. At timestep $\CT$ of the diffusion process, cross-attention maps and PFI are produced. HP network processes all inputs using specific feature extractors, and combines them using a Transformer-based decoder to predict whether generation should continue or not.}
    \label{fig:full_process_main_image}
    \vspace{-.3cm}
\end{figure*}

\tit{Text-Conditioning via Cross-Attention}
Stable Diffusion \citep{rombach2022high} employs cross-attention for text guidance. Indeed, the UNet incorporates self-attention followed by cross-attention layers, functioning across different resolutions (ranging from 64 to 8).  For each spatial dimension $P$ of the feature map and text tokens $N$ extracted from $y$, an attention map $A_t \in \mathbb{R}^{P \times P \times N}$ is formed, influencing the interaction between text tokens and spatial regions of the image.
The attention map extraction for a specific object $o$, given its index in the prompt, $i_o$, and the latent representation $z_t$ at time $t$, can be expressed as
\begin{equation}
\label{eq:attention_extraction}
A_{o,t} = \text{a}(i_o, z_t),
\end{equation}
where $A_{o,t}$ denotes the attention map associated with the object $o$ at timestep $t$, extracted from the latent representation $z_t$.

\section{Hallucination Early Detection+}
\label{sec:methods}

The primary goal of \HEAD is to detect and preemptively interrupt faulty generative processes. Our approach is designed to verify the consistency between the input textual guidance and the expected output \emph{during} the diffusion image generation process to save computational time and ensure correct generation. A distinctive feature of our approach is its capability to conduct this verification at a specific intermediate timestep of the diffusion pipeline.
To this aim, we designed an Hallucination Prediction (HP) network to detect the risk of hallucination. If an hallucination risk is predicted, our approach forces the generation process to restart using a different initial seed that might lead to better results.
In the following, we illustrate the proposed \HEAD approach at inference time to streamline the generation process and, as a result, enable automatic quality assessment of the final output.

\subsection{Multi-modal Inputs Extraction}
\label{subsec:input_extraction}
Consider a prompt $y$ that includes a set of target objects $O$ to be represented in the image. The procedure for identifying these target objects from the prompt is defined as follows:
\begin{equation}
\label{eq:TOE}
O = \text{TOE}(y),
\end{equation}
where $\text{TOE}(\cdot)$ denotes the Target Object Extraction function. In this context, the term ``objects'' pertains to words in the prompt that correspond directly to visible elements in the image. For these elements, the associated cross-attention maps will be derived. Although the current method is primarily focused on objects, it is designed to be extensible to encompass a broader range of visual concepts, such as attributes or colors, going beyond the limits of object-based extraction.

We then define the \textit{critical timestep}, denoted as $\CT$, as a specific step in the diffusion process where cross-attention maps for each object ($A_{O, \CT}$) and the Predicted Final Image (\PFItc) are extracted.

In particular, for each object $o \in O$, the cross-attention map $A_{o, \CT}$ is obtained by applying the function $a(\cdot)$, as described in Eq. \ref{eq:attention_extraction}. \PFItc, instead, represents the prediction of the expected outcome at the end of the generation process, using only information available at timestep $\CT$. In particular, the scheduler projects the latents at $\CT$ to the final step, and the decoder translates these predicted latents into the image space. Formally, this process can be defined as follows:
\begin{equation}
\begin{split}
\label{eq:pfi_projection}
\epsilon_{\CT} &= \epsilon_\theta(z_{\CT}, \CT) \\
z_{0}^{\CT} &= \Delta(z_{\CT}, \epsilon_{\CT}, \CT, 0) \\
\text{PFI}_{\CT} &= D(z_{0}^{\CT}),
\end{split}
\end{equation}
where $\epsilon_{\CT}$ represents the predictive noise obtained from the UNet model at critical timestep $\CT$. The function $\Delta$ updates the latents ${z}_{\CT}$ to the predicted latents at the final timestep, denoted as ${z}_{0}^{\CT}$. Finally, the decoder $D$ translates these predicted final latents into the Predicted Final Image, \PFItc, which represents a global snapshot at the given timestep and is utilized for predictions across all specified objects.

As an additional input for the model, we extract an embedding vector for each object. This vector is produced using CLIP~\citep{radford2021learning} from the text of the requested objects extracted from the prompt $y$. Specifically, for each object, we have $v_o = \text{CLIP}_{\text{Text}}(y_o)$ obtained by applying the CLIP model to the textual representation $y_o$ of each object.

\begin{figure*}[t]
    \centering
    \includegraphics[width=\linewidth]{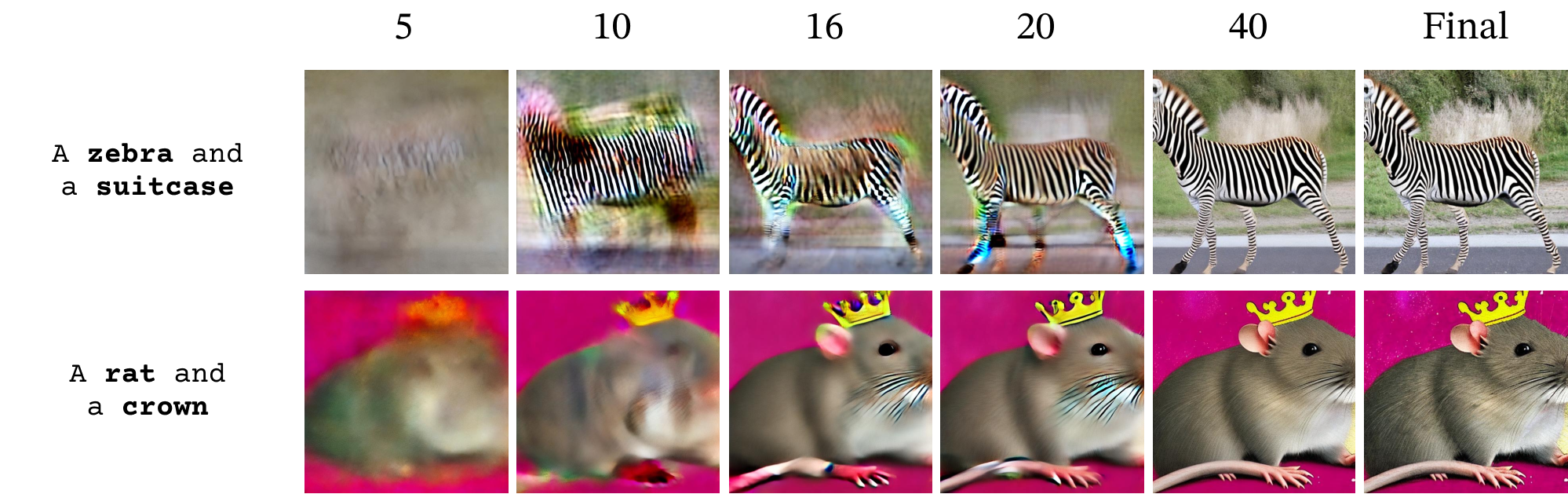}
    \vspace{-.1cm}
    \caption{Qualitative examples of the Predicted Final Image at different critical timesteps on simple prompts. Already from the 16\textsuperscript{th} step the final image is fully represented and the presence of objects can be predicted.}
    \label{fig:qualitativo_PFI}
    \vspace{-.3cm}
\end{figure*}

Examples of PFIs extracted at different timesteps are shown in Fig.~\ref{fig:qualitativo_PFI}. \PFItc, the attention maps $A_{O, \CT}$ and the textual embeddings $v_O$, enable the HP network to meticulously assess and predict the presence of specified objects in the final generated image, ensuring a coherent and accurate output aligned with the initial textual guidance.

\subsection{Hallucination Prediction Network}
\label{subsec:hallucination_prediction_network}

At a specific timestep $\CT$ of the generation, the Hallucination Prediction network evaluates each object $o \in O$ individually, utilizing the corresponding cross-attention map $A_{o, \CT}$, the textual embedding vector $v_o$, and the Predicted Final Image, \PFItc, as inputs. Subsequently, the network generates a binary output that indicates whether the target object is present or absent in the final image. Formally, the process can be written as
\begin{equation}
\label{eq:hallucination_prediction}
H_{o} = \text{HP}(A_{o}, \text{\PFItc}, v_o),
\end{equation}
where $H_{o}$ is the binary prediction for object $o$. An image is considered complete if $\forall o \in O, H_{o} = 1 $, otherwise, at inference time, the process must restart with a different seed.

The HP architecture consists of a Transformer model with cross-attention layers that integrate the three input streams (PFI$_\CT$, $A_{o, \CT}$, $v_o$) through self-attention and cross-attention mechanisms. Initially, all streams are elaborated separately. Indeed, the PFI is subjected to processing via the visual CLIP backbone, with the extraction of features from the last attention layer. This is followed by the concatenation of the textual token $v_o$ to this sequence. Simultaneously, a three-layer convolutional network, processes the cross-attention map.

Following these operations, both outputs are adjusted to the embedding dimension of the transformer model, predetermined at 192, by employing 1D convolution. This process sets the stage for the application of a self-attention mechanism on the stream derived from the CLIP processing This result (acting as the query) is integrated with a cross-attention layer with the features from the cross-attention maps (serving as keys and values). Skip connections are added following the architecture of the Transformer decoder~\citep{vaswani2017attention}. This attention block is consistently applied 12 times, concluding with a classification head that takes as input the activation corresponding to the first element of the input sequence, \ie~$v_o$. The model architecture is shown in Fig. \ref{fig:full_process_main_image}.

Throughout the training phase, the visual and textual backbones of CLIP are kept frozen. This approach guarantees substantial generalization capabilities towards objects not present in the training dataset, thereby accommodating applications on different datasets and scenarios.

\smallbreak
\subsection{Localization Module for Spatial Relations}
\label{subsec:localization_module}
\begin{revonly}
In addition to object-presence prediction, we introduce an integrated localization module that verifies pairwise spatial relations required by the prompt. This module shares the same backbone as the HP network and differs only in the final prediction head. It operates at the same critical timestep $\CT$ using the identical multi-modal inputs, namely $\text{PFI}_{\CT}$, cross-attention maps, and textual tokens.

\tinytit{Prompt relation extraction} Let $O$ be the set of objects extracted as in Eq.~\ref{eq:TOE}. If the prompt specifies positional constraints between two objects, we define the set of relations using a compact extractor. Let $\Pi=\{\text{top},\text{bottom},\text{left},\text{right}\}$. We compute
\begin{equation}
\label{eq:relations}
\mathcal{R} = \text{REL}(y, O) \subseteq O \times O \times \Pi,
\end{equation}
where $\text{REL}(\cdot)$ is a lightweight LLM-based relation extractor, applied in parallel with the early diffusion steps, similarly to object extraction described in Sec.~\ref{subsec:training_object_extraction}.

\tinytit{Centroid estimation} For each object $o \in O$, the Localization Prediction (LP) network predicts the centroid $c_o = (x_o, y_o)$ of its final placement in image coordinates using inputs at $\CT$. The centroid is obtained by a regression head over the shared features computed from $\text{PFI}_{\CT}$, the cross-attention map $A_{o,\CT}$, and the text token $v_o$:
\begin{equation}
\label{eq:centroid_pred}
c_o = \text{LP}(A_{o,\CT}, \text{PFI}_{\CT}, v_o),
\end{equation}
where $\text{LP}(\cdot)$ shares the backbone of HP and replaces the final classification layer with a 2D regression layer for $(x_o,y_o)$.

\tinytit{Relation checking} Given two objects $i$ and $j$, their predicted centroids $c_i=(x_i,y_i)$ and $c_j=(x_j,y_j)$ are compared to verify whether each relation in $\Pi$ holds (e.g., above/below or left/right) using straightforward coordinate comparisons with a small tolerance (5\% of image size).
For a relation $r=(i,j,\rho) \in \mathcal{R}$, the localization module produces a binary decision
\begin{equation}
\label{eq:relation_check}
L_{i,j,\rho} = \mathbbm{1}[\, \rho(i,j) \text{ holds under } c_i, c_j \,],
\end{equation}
which evaluates whether the positional constraint is satisfied at $\CT$.

\tinytit{Joint gating with HP} The overall decision to continue generation at timestep $\CT$ is the conjunction of module decisions. Let $H_o$ be the per-object presence predictions from Eq.~\ref{eq:hallucination_prediction}. If no spatial relations are specified ($\mathcal{R}=\emptyset$), we recover the original criterion $\forall o \in O,\, H_o=1$. Otherwise, we require
\begin{equation}
\label{eq:gating}
G = \Big( \forall o \in O,\, H_o=1 \Big) \wedge \Big( \forall (i,j,\rho) \in \mathcal{R},\, L_{i,j,\rho}=1 \Big).
\end{equation}
Generation proceeds if and only if $G=1$; otherwise, the current run is aborted and restarted with a new seed, consistent with the \HEAD policy. This module is model-agnostic and can be enabled alongside other modules within the same gating scheme.
\end{revonly}

\section{InsideGen Dataset}
\label{sec:training}

\subsection{Prompts Selection}
\label{subsec:dataset_selection}
To train the HP network, we require a dataset containing intermediate outputs from the diffusion process. Since our goal is to train \HEAD on realistic scenarios, we curated 4,100 prompts from the dataset introduced by~\cite{bakr2023hrs}, ensuring coverage of diverse objects and situations. These prompts are built specifically to stress different generation capabilities, namely spatial composition, size composition, action composition, fairness, and bias. To augment the dimensionality of the dataset and thoroughly investigate output variations influenced by different seeds, we generated 11 images for each prompt using distinct seeds. In total, we generated nearly 45,000 images, both with Stable Diffusion v1.4 and Stable Diffusion v2.0 generators~\citep{rombach2022high}.
During generation, we fixed 50 steps of the diffusion process and we collected the PFI and cross-attention maps $A$ at multiple time steps for every requested objects to give the possibility to make different tests for different critical timesteps $\CT$\footnote{In particular, the critical steps $\CT$ are chosen as follows: [0, 1, 2, 3, 4, 5, 6, 7, 8, 9, 10, 12, 14, 16, 18, 20, 25, 40]}. Dataset was divided in train, val and test (80-10-10), splitting over the prompts, keeping all seeds in the same set.

\subsection{Target Objects Extraction}
\label{subsec:training_object_extraction}
During the dataset creation process and during real-time inference, objects, denoted as $O$, are extracted from the textual prompt \( y \), as described in Eq.~\ref{eq:TOE}. For this purpose, we employed GPT-3.5~\citep{openai_gpt-4_2023}, selected for its robust zero-shot generalization abilities. This method stands in contrast to conventional text tagging techniques that generally necessitate specific training for each domain.

The extraction procedure is time-efficient and can be executed concurrently with the initial diffusion steps. This parallel processing is made possible by selecting a non-initial critical timestep \( \CT \), which enables the object extraction to proceed in tandem with the diffusion process.

We instructed the system to use a specific prompt to guide its entity recognition process. The prompt used was as follows:

\begin{minipage}{\linewidth}
\small
\texttt{\\You are a system that is able to recognize entities in a text. \\Entities are objects, people, animals, etc. that have a physical representation. Avoid to include abstract subjects. Do not consider adjectives in the entities.\\}
\end{minipage}

To enhance the model accuracy, we also provided a few-shot learning approach with relevant examples. This method was crucial in ensuring that the model was focused on extracting only concrete entities while excluding abstract concepts and adjectives, aligning with the objectives of our research and the operational requirements of the \HEAD pipeline.

\subsection{Automatic Label Creation}
\label{subsec:label_creation}

A key aspect of our dataset collection approach is an automatic labeling pipeline to verify the presence of specific objects in the generated images. A strong requirement for this labeling system is the ability to operate without a predefined set of object labels, necessitating the use of an open vocabulary approach. This approach provides flexibility and adaptability in identifying a wide range of objects, regardless of their characteristics. According to this objective, we adopted OWLv2~\citep{minderer2023scaling}, a open vocabulary detector renowned for its robust detection capabilities and for providing confidence scores for each identified object. 
\rev{Furthermore, we leverage OWLv2 to extract object centroids, which are crucial for training the LP network.}
The ability of this model to accurately detect and quantify the presence of various objects significantly improved the reliability and integrity of our dataset.
The likelihood of achieving a \perfect generation, meaning all requested objects are present in the final image, for each object count and model type is depicted in Fig. \ref{fig:first_page}(a).

\begin{figure*}[t]
    \centering
    \includegraphics[width=0.8\linewidth]{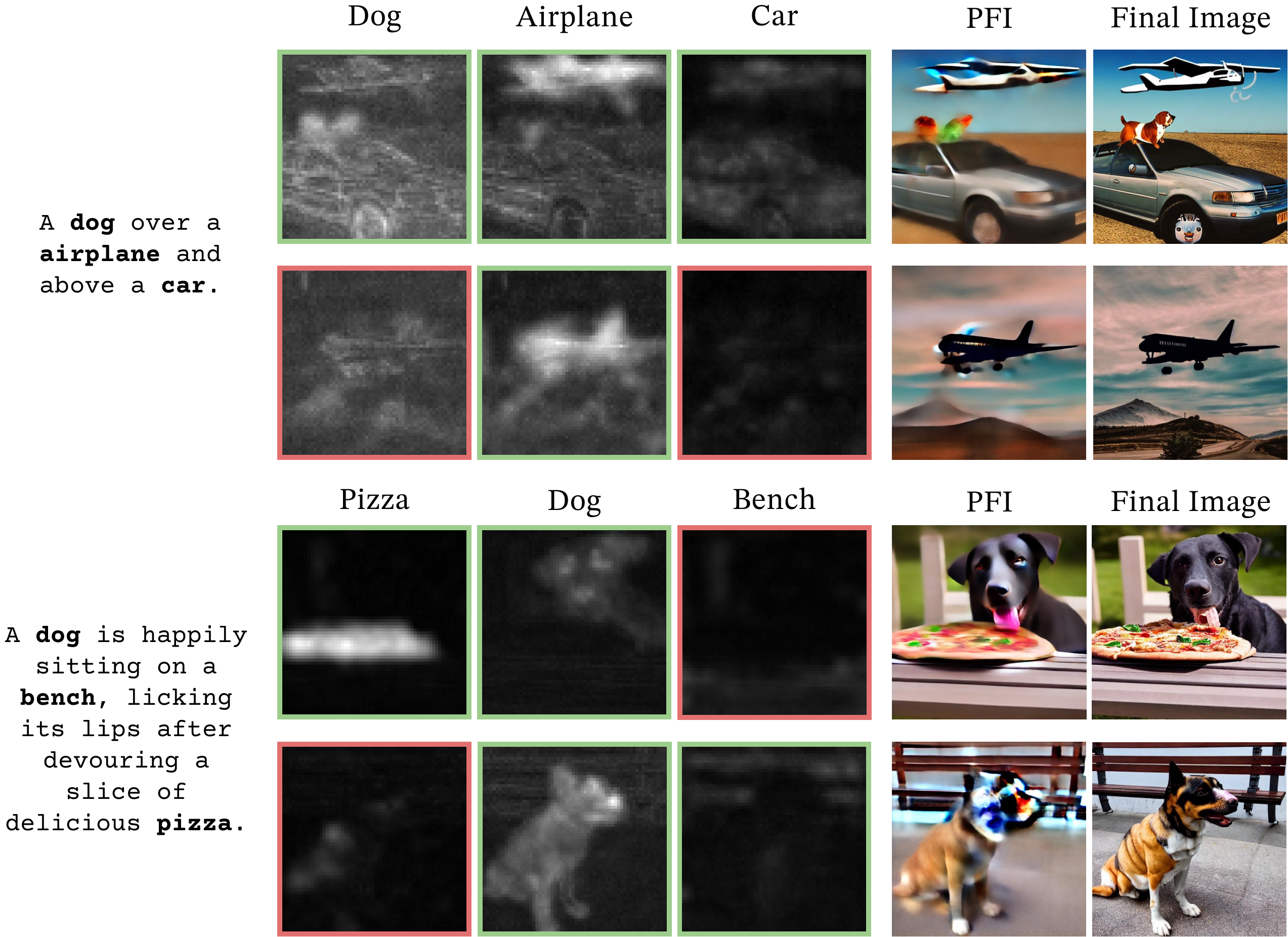}
    \vspace{0.25cm}
    \caption{Examples of Target Objects Extraction, cross-attention maps, and the Predicted Final Image at timestep $\CT=16$. The cross-attention maps are highlighted with a green border when they correspond with the object in the image; otherwise, they are highlighted in red.}
    \label{fig:suppqualitatives}
\end{figure*}

\subsection{\DATASETNAME qualitatives}
We present different qualitative samples of the \DATASETNAME dataset in Figure~\ref{fig:suppqualitatives}.
In these examples, we performed the object extraction pipeline following the procedure detailed in Section~\ref{subsec:training_object_extraction}, and generated the images using Stable Diffusion 1.4~\citep{rombach2022high}.
Notably, first insights on subject hallucinations are still detectable at timestep 16 of the generation process. For instance, considering the prompt \texttt{A dog over a airplane
and above a car}, the second row does not represent either the \texttt{dog} or the \texttt{car} in its PFI. Moreover, the cross-attention maps of these missing subjects are less emphasized compared to the upper row, where all the objects are well represented.
Similar outcomes are observed in the prompt \texttt{A dog is happily sitting on a bench, licking its lips after devouring a slice of delicious pizza}. Indeed, \texttt{pizza} is missing from both the final image and the PFI in the example in the fourth row. Compared to the third instance, where all the subjects are well-represented in the PFI, the cross-attention map is more activated in the case of \texttt{pizza} subject.

\section{Experiments}
\label{sec:results}
\tit{Training and implementation details} The training for the Hallucination Prediction network is conducted using a learning rate of $5 \times 10^{-4}$, with a reduction on plateau strategy. Specifically, the learning rate is reduced by a factor of 0.5 if there is no improvement observed for 10 epochs, demonstrating a patient approach to convergence. HP is trained on SD2 with a fixed batch size of 128. From a technical perspective, the Transformer Decoder is equipped with 12 self-attention and cross-attention blocks, each of which has 3 attention heads. The convolutional layers that process the cross-attention maps instead have a $3\times3$ kernel with stride 1 and padding 1 with an output channel of 32, 64, and 128 for respectively the first, second, and third layers. Following each convolutional layer, batch normalization, ReLU activation, and MaxPooling operations are sequentially applied. An exception resides in the terminal convolutional layer, which incorporates adaptive average pooling to refine the dimensionality to a $14\times14$ feature map. This resultant map is then flattened and projected to the 192 embedding dimension of the Transformer decoder.

\begin{table}[t]
  \centering
  \caption{TN-Rate, Recall, and percentage of time saved for the \HEAD model at different timesteps. In addition to the model name, the timestep $\CT$ used for training the model is specified. \OLDHEAD refers to our previous method~\citep{betti2024optimizing}}

  \begin{tabular}{lccc}
    \toprule
    Model & \textbf{TN-Rate} & \textbf{Recall} & \textbf{Time Saved} \\
    \midrule
    \HEAD 5 & 50.16 & 90.22 & \textbf{37.73} \\
    \HEAD 8 & 56.93 & 91.06 & 36.37 \\
    \HEAD 10 & 58.37 & 90.09 & 30.03 \\
    \HEAD 12 & 63.36 & 91.15 & 30.11 \\
    \HEAD 14 & 61.65 & 93.42 & 32.91 \\
    \HEAD 16 & 64.42 & 93.60 & 30.03 \\
    \HEAD 18 & 66.86 & 93.13 & 25.64 \\
    \HEAD 20 & 65.43 & \textbf{94.81} & 26.85 \\
    \HEAD 25 & \textbf{76.95} & 93.40 & 14.18 \\
    \midrule
    \OLDHEAD 8 & 43.73 & 85.67 & 11.55 \\
    \OLDHEAD 25 & 52.16 & 88.02 & -11.70 \\
    \bottomrule
  \end{tabular}

  \vspace{.2cm}
  \label{tab:first_table}
\end{table}

\subsection{Time Saving Experiment}
\label{sec:time_saving}

This experiment is designed to evaluate a specific aspect of the performance of our method, namely its ability to achieve a \perfect generation. Traditional models typically require a full run followed by a post-generation evaluation, restarting the process if the output is not \perfect. The unique advantage of employing \HEAD lies in its capability to perform evaluations during the generation process itself, potentially allowing for much earlier restarts when non-ideal outputs are detected. This can significantly reduce the overall time spent on generating and evaluating images. It is important to note that, for the purposes of this experiment, we do not account for the time that would normally be consumed by a model without \HEAD in conducting post-generation evaluations.

\tit{HP Performance Impact}
The HP network predicts whether to continue or halt the image generation process. When a True Positive (TP) occurs, the correct generation proceeds uninterrupted, having no effect on computation time. Conversely, a False Positive (FP) allows an incorrect generation to continue without interruption, thus missing an opportunity for time savings, but still not impacting computation time. A True Negative (TN) indicates an incorrect generation has been correctly halted, leading to time savings. Finally, a False Negative (FN) means a correct generation is mistakenly stopped, resulting in a loss of time.

The HP network is expected to demonstrate proficient recall performance, characterized by minimal FNs, thereby preventing the unwarranted cessation of correct runs. Concurrently, the network should achieve high true-negative rate, characterized by a minimal number of FPs relative to TNs. This is critical to ensure the timely interruption of incorrect generations, thereby enhancing time-saving. Consequently, in the training phases of the HP network, emphasis was placed on achieving both recall and true-negative rate metrics.

\tit{Best $\CT$ selection}
The most impactful hyperparameter in \HEAD is $\CT$, the step in which the HP network is applied. An earlier detection, with a small $\CT$, can potentially lead to greater time savings in the case of TNs (correct hallucination identification).  Nevertheless, this approach is met with a notable challenge: during the preliminary phases, the quality of cross-attention maps and PFIs is suboptimal, thereby affecting the performance efficacy of the HP network.

To quantify the time saved or lost using the model, we conducted Monte Carlo simulations until a \perfect generation is found. The detailed algorithm and pseudo-code are provided in the Appendix. We compared the same model trained on different $\CT$ and results are shown in Table \ref{tab:first_table}.
Recall values are maintained using smaller $\CT$ values while TN-Rate changes significantly. The overall time saved for a \perfect generation in an ideal scenario with \HEAD used with $\CT=5$ is the highest, with a potential 37.73\% of time saved over a normal version of SD2. Early detection has the biggest impact considering recall values are similar.

We compared the current models with \OLDHEAD~\citep{betti2024optimizing} models which exhibit significantly poorer performance on the \DATASETNAME dataset. For instance, when comparing at the same timestep $\CT=25$, \HEAD increases time-saving by 25.88\%. Further, \OLDHEAD obtains no benefit (-11.70\%) when introduced in this complex scenario. Similarly when comparing the methodologies at lower timestep $\CT=8$, \HEAD obtains a gain of 26.18\% in time-saving.

\begin{table*}[t]
\caption{Comparison with state-of-the-art methodologies, on the evaluation protocol of~\cite{wang2023tokencompose}. Results without \HEAD are taken from \cite{wang2023tokencompose}.}
  \label{tab:method_performance_coco}
  \centering
  \setlength{\tabcolsep}{1.25em} 
    \small
    \begin{tabular}{lcccc}
      \toprule
       & \multicolumn{4}{c}{COCO} \\
      \cmidrule{2-5}
        Method & MG2 & MG3 & MG4 & MG5 \\
      \midrule
      SD 1.4~\citep{rombach2022high} & 90.72\textsubscript{1.33} & 50.74\textsubscript{0.89} & 11.68\textsubscript{0.45} & 0.88\textsubscript{0.21} \\
      \rowcolor{LightCyan}
      SD 1.4 w/ \HEAD & \textbf{95.44}\textsubscript{0.67} & \textbf{65.03}\textsubscript{1.76} & \textbf{19.53}\textsubscript{1.73} & \textbf{1.78}\textsubscript{0.24} \\
      \midrule
      Composable Diffusion~\citep{liu2022compositional} & 63.33\textsubscript{0.59} & 21.87\textsubscript{1.01} & 3.25\textsubscript{0.45} & 0.23\textsubscript{0.18} \\
      Layout Guidance~\citep{chen2024training} & 93.22\textsubscript{0.69} & 60.15\textsubscript{1.58} & 19.49\textsubscript{0.88} & 2.27\textsubscript{0.44} \\
      Structured Diffusion~\citep{feng2023trainingfree} & 90.40\textsubscript{1.06} & 48.64\textsubscript{1.32} & 10.71\textsubscript{0.92} & 0.68\textsubscript{0.25} \\
      Attend-and-Excite~\citep{chefer2023attendandexcite} & 93.64\textsubscript{0.76} & 65.10\textsubscript{1.24} & 28.01\textsubscript{0.90} & 6.01\textsubscript{0.61} \\
      Token Compose~\citep{wang2023tokencompose} & 98.08\textsubscript{0.40} & 76.16\textsubscript{1.04} & 28.81\textsubscript{0.95} & 3.28\textsubscript{0.48} \\
      \rowcolor{LightCyan}
     Token Compose w/ \HEAD & 97.61\textsubscript{0.40} & 81.27\textsubscript{1.40} & 35.33\textsubscript{1.97} & 4.93\textsubscript{0.57} \\
     \midrule
     \rev{PixArt-$\alpha$~\citep{chen2024pixartalpha}}   & \rev{98.59\textsubscript{0.49}}          & \rev{83.19\textsubscript{2.18}}          & \rev{42.80\textsubscript{4.09}}          & \rev{8.80\textsubscript{2.10}}           \\
     \rowcolor{LightCyan}
     \rev{PixArt-$\alpha$ w/ \HEAD}                      & \rev{\textbf{99.25}\textsubscript{0.28}} & \rev{\textbf{89.19}\textsubscript{0.77}} & \rev{\textbf{51.65}\textsubscript{1.89}} & \rev{\textbf{12.72}\textsubscript{1.46}} \\
     \bottomrule
    \end{tabular}

\end{table*}

\subsection{Generation Quality Comparison}
To compare the final image quality after a full generation, we followed the evaluation protocol introduced in~\cite{wang2023tokencompose}. Specifically, in this benchmark, the text-to-image model is prompted to generate five subjects (\ie~\textit{A, B, C, D, E}) with the caption \textit{A photo of A, B, C, D, and E}. Subsequently, an open-vocabulary object detector~\citep{minderer2023scaling} is asked to detect the presence of the subjects in the generated images.
In Table \ref{tab:method_performance_coco} the performance of different models, in their base form or with \HEAD, are presented. This split includes 80 objects extracted from COCO~\citep{lin2014microsoft} combined as previously defined, building 1000 different prompts.
Considering the capability of each model to potentially introduce hallucinations, the generated images may feature between one to five subjects. The metric MG-$N$ is utilized to quantitatively ascertain the count of generated images that accurately portray a minimum of $N$ subjects as requested in the initial prompt. Results are reported with mean and standard deviation over 10 seeds following the original implementation.

To promote a fair comparison, we impose a limitation on the maximum number of restarts for \HEAD, specifically setting this limit to five iterations. This constraint is designed to standardize the inference time, facilitating a direct comparison with alternative methodologies. In instances where a \perfect image fails to materialize within these iterations, the seed with the highest number of objects predicted by \HEAD is chosen for the generation.
Further, the \HEAD model is trained on SD2 on \DATASETNAME, which differs significantly from the datasets used in these experiments.
Additionally, the critical timestep $\CT$ is set to 25.
\rev{The number of inference steps is set to 50 for SD models and 20 for PixArt, maintaining the default settings; similarly, the critical timestep $\CT$ is set to 25 and 10 for SD models and PixArt, respectively.}
The defined \HEAD methodology proves beneficial in enhancing the overall quality of the output.

As illustrated in Table~\ref{tab:method_performance_coco}, the adoption of \HEAD enhances the detection of object presence in comparison to SD 1.4~\citep{rombach2022high}. Specifically, an average increase of 14.29\% in the detection of three objects is observed when \HEAD is integrated with SD1.4 compared to the raw SD. Additionally, improvements of 7.85\% and 0.9\% are noted for MG4 and MG5, respectively.
When compared to Composable Diffusion~\citep{liu2022compositional}, Layout Guidance~\citep{chen2024training}, and Structured Diffusion~\citep{feng2023trainingfree}, \HEAD with SD1.4 surpasses them in nearly all the MG categories. Specifically, for MG3 metric, SD1.4 equipped with \HEAD obtains a gain of 43.16\%, 4.8\%, and 16.39\% over Composable, Layout Guidance, and Structured Diffusion respectively.

Significantly, \HEAD was initially trained using the model SD2, which demonstrates the capability of the model to adapt to various diffusion models beyond the one it was originally trained on. 
\rev{This adaptability is further corroborated by performance gains when \HEAD is used with TokenCompose~\citep{wang2023tokencompose}: \HEAD yields improvements of 5.11\%, 6.52\%, and 1.65\% on MG3--MG5 over TokenCompose. Moreover, with Transformer-based PixArt-$\alpha$~\citep{chen2024pixartalpha}, \HEAD achieves the best MG scores across MG2--MG5 (Table~\ref{tab:method_performance_coco}). 
Overall, these results confirm that \HEAD is model-agnostic, effectively complementing both UNet- and Transformer-based diffusion generators without retraining the target model.}

\subsection{Ablation on Input Types}

\begin{table}[t]
\vspace{-.7cm}
  \centering
  \caption{Ablation of different input types: TN-Rate and Recall for different model settings. Experiments done at $\CT=25$. In the Table XA and Text represent cross-attention maps and $v_o$, respectively. \HEAD is the model with all input types that we have used throughout our experiments.}
  \small 
  \setlength{\tabcolsep}{2em} 
  \begin{tabular}{lcc}
    \toprule
    \textbf{Name} & \textbf{TN-Rate} & \textbf{Recall} \\
    \midrule
    Text + XA & 62.35 & 93.82 \\
    Text + PFI & 74.84 & 91.21 \\
    PFI + XA & 33.72 & \textbf{95.22} \\
    \HEAD & \textbf{76.95} & 93.40 \\
    \bottomrule
  \end{tabular}
  \vspace{.2cm}
  \label{tab:ablation_study}
\end{table}

To understand the individual impact of cross-attention maps, PFI, and the textual feature vector on the final assessment, we conducted an ablation study. This involved creating variations of the HP network, each excluding one of the inputs, and comparing their performance.
The analysis reveals that the most accurate results are achieved when all inputs are incorporated. Specifically, when the textual token $v_o$ is dropped, the performance decreases by 24\% in TN-Rate. The strong interaction between CLIP features extracted from the PFI and $v_o$ is relevant for \HEAD, as it precisely specifies the target elements that the model needs to identify within the visual inputs. Indeed, when only textual information and cross-attention maps are combined performance still takes a decrease of 12.41\% on TN-Rate, proving the coupling between the textual input and PFI. Although the cross-attention map has the least impact compared to the other inputs, it increases Recall by 1.19\% and TN-Rate by 2.11\% compared to the model without it. This is attributable to the ability of the cross-attention map to guide the focus of the model toward certain regions within the image, facilitating the verification of object presence.

\subsection{Qualitative Comparison}

\begin{figure*}[t]
    \centering
    \includegraphics[width=0.8\linewidth]{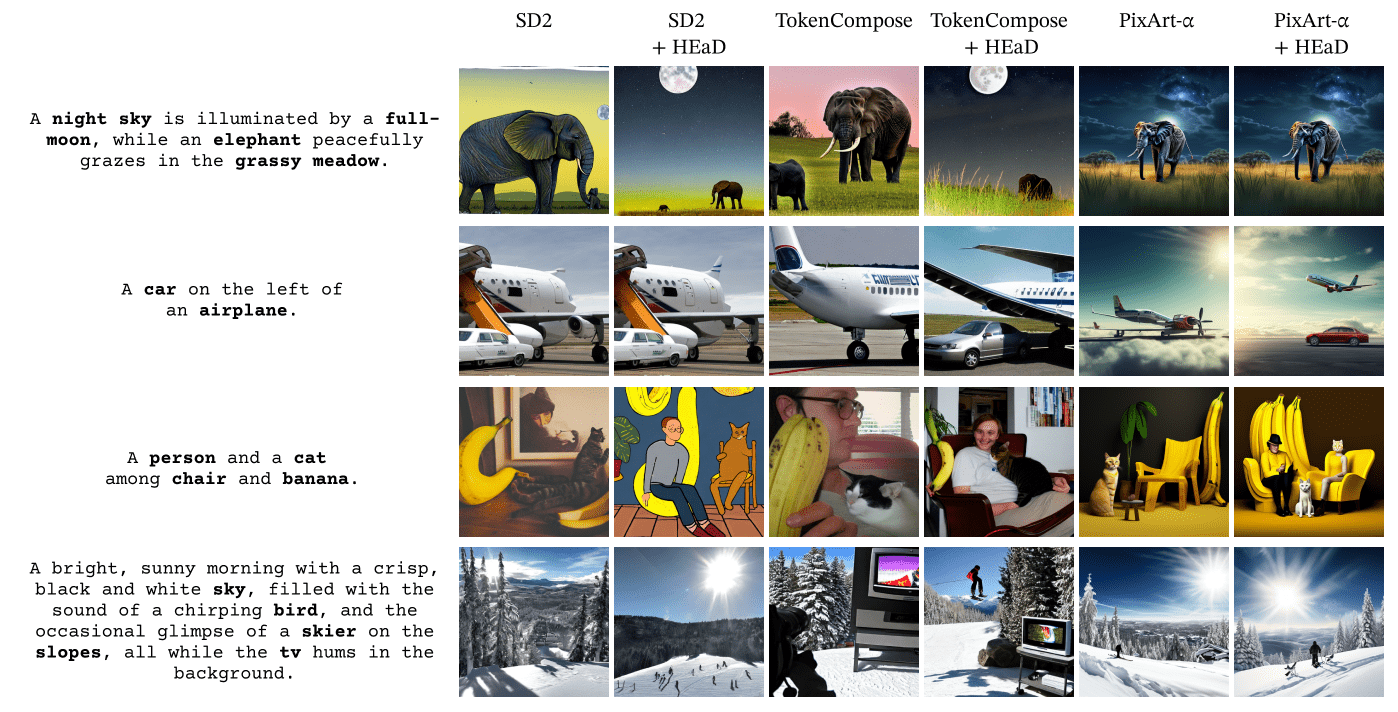}
    \caption{Qualitative analysis on the comparison between a normal Diffusion Model and the same version using \HEAD. The same starting seed was used for all the experiments on the same prompt.}
    \label{fig:qualitativi_head}
    \vspace{-.3cm}
\end{figure*}

In Figure~\ref{fig:qualitativi_head}, we provide some qualitative comparisons of generated images by SD2 and TokenCompose with and without \HEAD, evaluated on prompts from \DATASETNAME test set.
The four rows represent different contributes of \HEAD. In the top row, both SD2 and TokenCompose fail to represent correctly the subject \texttt{moon}. Triggered by this hallucination, \HEAD reiterates with another seed until a \perfect generation is reached. Conversely, in the second row, SD2 accurately renders both \texttt{car} and \texttt{aircraft}, inducing \HEAD to allow the generative process to continue, thereby yielding an identical result. In contrast, TokenCompose exhibits a failure by negating the presence of the \texttt{car} during its generative process, a deviation promptly addressed by \HEAD through the termination of generation and the introduction of a new seed. Similarly, in the third row, SD2 hallucinates \texttt{person} and TokenCompose \texttt{chair} while \HEAD corrects them. The concluding row presents a scenario very complex with five objects in the prompt; SD2 solely visualizes \texttt{slopes}, and \HEAD, despite its efforts, fails to generate all intended objects, reaching the maximum number of iterations. Nonetheless, in its interaction with TokenCompose, \HEAD successfully intervenes by incorporating the omitted \texttt{skier}, showcasing its ability to correct hallucinations effectively.

\subsection{Localization Module Results}
\label{subsec:loc_results}
\begin{revonly}
\tinytit{Training} We train the Localization Prediction (LP) network with the same optimization setup described in Section~\ref{sec:results}, using object centroids from \DATASETNAME for non-hallucinated objects.

\tinytit{Evaluation protocol} To validate the LP network, we sample two objects from COCO and compose a two-object prompt with a single positional relation $\rho\in\Pi=\{\text{top},\text{bottom},\text{left},\text{right}\}$. For each prompt, we generate 5 images with different seeds and collect inputs at the critical timestep $\CT$, exactly as for HP.
Following the procedure in Section~\ref{sec:results}, we evaluate three model configurations: the base generator, the base generator with \HEAD, and the base generator augmented with both \HEAD and the LP network. We verify the positional relation on generated images by applying OWLv2 to detect object centroids and checking whether $\rho$ holds within tolerance of a 5\% of the image size. 
Subsequently, we report the following metrics: (i) MG2: the percentage of images in which both objects are present; (ii) MG$_{\text{loc}}$: the percentage of images in which both objects are present and the specified positional relation is satisfied; and (iii) Relation Consistency: the proportion of correct relations among all images in which both objects are present.

\tinytit{Discussion}
Table~\ref{tab:localization_results} presents the evaluation of the localization module. Consistent with prior findings, incorporating \HEAD yields systematic improvements in MG2 over the base models. Furthermore, enabling the localization module brings additional gains in both MG$_{\text{loc}}$ and Relation Consistency, as generations predicted to violate the specified positional relation are terminated early and resampled. For example, Relation Consistency increases by 22.0\%, 13.9\%, and 17.5\% relative to SD 1.4, SD 2, and TokenCompose with \HEAD but without the localization module, respectively. 
Relative to the base models, the most substantial improvement is observed with TokenCompose: equipping it with \HEAD and the localization module raises Relation Consistency from 48.5\% to 67.5\%, indicating that relation-consistent generations are encouraged more effectively than with either the base model alone or the base model plus \HEAD.
\end{revonly}

\rev{
\begin{table*}[t]
  \caption{\rev{Impact of the localization module. MG2: both objects present. MG$_{\text{loc}}$: both objects present and the required relation satisfied. Relation Consistency (\%) is the percentage of correct relations among images where both objects are present.}}
  \label{tab:localization_results}
  \centering
  \setlength{\tabcolsep}{1.25em}
  \small
  \begin{tabular}{lccc}
    \toprule
    \rev{Method}                                  & \rev{MG2}           & \rev{MG$_{\text{loc}}$} & \rev{Relation Consistency (\%)} \\
    \midrule
    \rev{SD 1.4}                                  & \rev{39.6}          & \rev{19.2}              & \rev{48.5}                      \\
    \rev{SD 1.4  w/ \HEAD wo/ localization}       & \rev{60.2}          & \rev{20.0}              & \rev{33.2}                      \\
    \rowcolor{LightCyan}
    \rev{SD 1.4 w/ \HEAD w/ localization}         & \rev{\textbf{61.2}} & \rev{\textbf{33.8}}     & \rev{\textbf{55.2}}             \\
    \midrule
    \rev{SD 2}                                    & \rev{51.6}          & \rev{22.2}              & \rev{43.0}                      \\
    \rev{SD 2  w/ \HEAD wo/ localization}         & \rev{\textbf{69.2}} & \rev{29.0}              & \rev{41.9}                      \\
    \rowcolor{LightCyan}
    \rev{SD 2 w/ \HEAD w/ localization}           & \rev{68.8}          & \rev{\textbf{38.4}}     & \rev{\textbf{55.8}}             \\
    \midrule
    \rev{Token Compose}                           & \rev{67.6}          & \rev{32.8}              & \rev{48.5}                      \\
    \rev{Token Compose w/ \HEAD wo/ localization} & \rev{\textbf{71.3}} & \rev{35.7}              & \rev{50.0}                      \\
    \rowcolor{LightCyan}
    \rev{Token Compose w/ \HEAD w/ localization}  & \rev{70.8}          & \rev{\textbf{47.8}}     & \rev{\textbf{67.5}}             \\
    \bottomrule
  \end{tabular}
\end{table*}
}

\section{Limitations}
In this paper we introduced \HEAD, a model designed to prevent the generation of hallucinated images with missing subjects. Nevertheless, it is critical to acknowledge that the model's capability to predict hallucinations is not flawless. In certain instances, it may fail to accurately capture all subjects within an image or may prematurely halt the generation process of an image that could have been depicted correctly if allowed to complete. Moreover, while the primary focus of \HEAD is on the absence of subjects, other types of inconsistencies may also manifest during image generation. For example, text-to-image models might inaccurately portray spatial relationships between objects~\citep{chen2024training} or fail in the linguistic binding of entities and modifiers~\citep{rassin2023linguistic}.

\section{Conclusions}
\label{sec:conclusions}

This paper presents \HEAD, a state-of-the-art methodology designed to significantly improve the efficiency and accuracy of image generation processes using diffusion models. The key to our innovation lies in the integrated use of cross-attention maps, Predicted Final Image, and textual data to predict the outcome of the image generation process. \HEAD has been shown to improve the generation fidelity of the requested objects inside the final image.
A crucial takeaway from our research is that \HEAD, in its current form without retraining, can be seamlessly integrated with various diffusion models, reliably ensuring that all specified objects are accurately represented in the final image.
The internal operations of the diffusion model are not affected by the functionality of \HEAD.
Furthermore, we created \DATASETNAME, which is a dataset that consists of intermediate diffusion output (attention maps, and PFIs) with annotated hallucinations. This resource will facilitate further investigation into how the internal generation data can be leveraged to refine the image generation process as a whole.


\bibliography{sn-bibliography}

\newpage
\begin{appendices}

\section{Monte Carlo HEaD+ simulation}
The Python pseudocode detailed in Algorithm~\ref{algo:monte_carlo_pseduocode} simulates the time savings achieved by implementing the \HEAD approach within the image generation process. Its effectiveness depends on the performance of the model, particularly in terms of recall, true-negative rate, and the number of subjects present $|O|$. \HEAD analyzes each subject independently, and it only requires one of the objects to be predicted as absent to halt the generation and restart with a new seed. The time saving occurs when the model incorrectly generates an image, \ie~a subject is not present, and \HEAD is able to predict this and immediately restart the generation with a different seed. The time saved in each of these instances is dependent on $\CT$, which represents the timestep in which the prediction is done.

\lstset{
    language=Python,
    basicstyle=\ttfamily\small,
    keywordstyle=\color{blue},
    stringstyle=\color{red},
    commentstyle=\color{green!60!black},
    numbers=left,
    numberstyle=\tiny\color{gray},
    stepnumber=1,
    numbersep=10pt,
    tabsize=4,
    showstringspaces=false,
    breaklines=true,
    breakatwhitespace=true,
    frame=single,
    captionpos=b,
    morekeywords={self} 
}

\onecolumn
\begin{lstlisting}[caption={Python pseudo code for \HEAD Monte Carlo simlulation.}, label={algo:monte_carlo_pseduocode}]
# complete_generation_probability: probability of complete image
# recall: recall of the HP network
# TN_Rate: TN_Rate of the HP network
# time_per_model_iteration: time for completing a generation
# max_step_used: last step used for HEaD evaluation
# num_objects: number of objects to evaluate
# total_steps: number of generation step, 50 for SD2
# num_simulations: number of Monte Carlo simulations

# Computing time when HEaD model detects failure
time_used_per_TN = (max_step_used / total_steps) * time_per_model_iteration
# Time with HEaD approach
time_with_head = 0

for _ in range(num_simulations):
    success = False
    while not success:
        # Generate an image that will be complete
        # with a probability of complete_generation_probability
        actual_success = random() < complete_generation_probability
        if actual_success:
            # HP network must predict all 1s to halt the generation
            model_predicts_success = all(
                random.random() < recall for _ in range(num_objects)
            )
            if model_predicts_success: # TP
                time_with_head += time_per_model_iteration
                success = True
            else: # FN
                time_with_head += time_per_model_iteration
        else:
            # The generation has at least one object hallucinated.
            # The HP must find at least one hallucinated object to
            # restart the generative process
            model_predicts_failure = any(
                random() < TN_Rate
                for _ in range(num_objects)
            )
            if model_predicts_failure: # TN
                time_with_head += time_used_per_TN
            else: # FP
                time_with_head += time_per_model_iteration

# Time with HEaD approach
avg_time_with_HEaD = time_with_head / num_simulations

# Time without HEaD approach
avg_time_no_HEaD = time_per_model_iteration / complete_generation_probability

return 1 - avg_time_with_HEaD / avg_time_no_HEaD
\end{lstlisting}
\twocolumn




\section{Additional qualitative comparisons}\label{secA2}
\begin{figure*}[t]
    \centering
    \includegraphics[width=\linewidth]{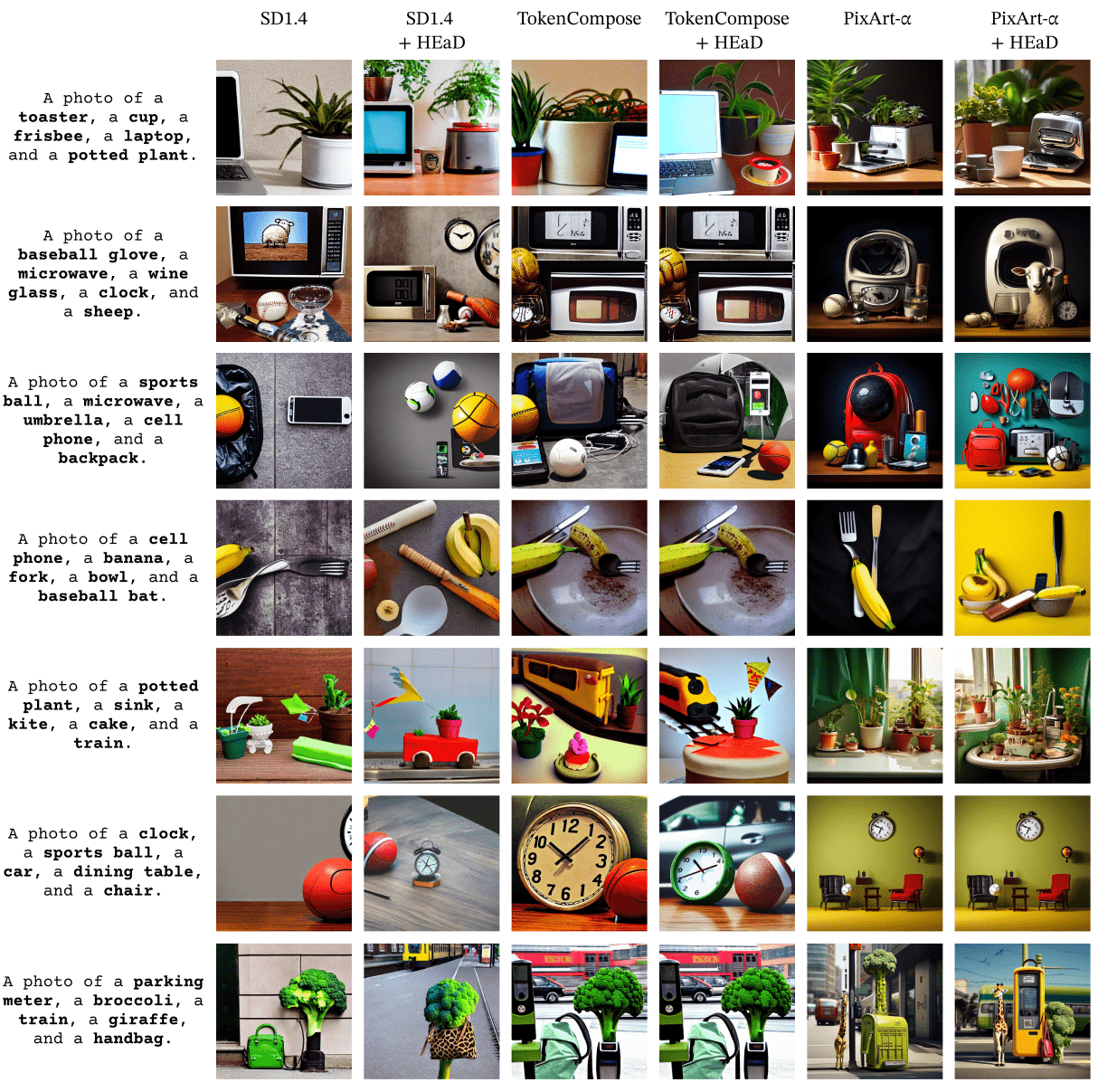}
    \caption{Extended qualitative comparison across SD~1.4, TokenCompose, and PixArt-$\alpha$ with and without \HEAD on multi-object prompts. \HEAD helps recover missing subjects and yields images that better satisfy the requested compositions.}
    \label{fig:appendix_qualitatives_big}
\end{figure*}

In addition to the main qualitative analysis in Fig.~\ref{fig:qualitativi_head}, we provide a larger panel of comparisons between base generators and their counterparts equipped with \HEAD. As illustrated in Fig.~\ref{fig:appendix_qualitatives_big}, \HEAD systematically recovers missing subjects and improves compositional correctness across SD~1.4, TokenCompose~\citep{wang2023tokencompose}, and PixArt-$\alpha$~\citep{chen2024pixartalpha}. These examples complement the quantitative gains discussed in Section~\ref{sec:results} and in Table~\ref{tab:method_performance_coco}.

\end{appendices}

\end{document}